\documentclass[12pt]{article}
\usepackage{hyperref}
\usepackage{arxiv}
\usepackage{color,soul}

\usepackage[utf8]{inputenc} 
\usepackage[T1]{fontenc}    
\usepackage{hyperref}       
\usepackage{url}            
\usepackage{booktabs}       
\usepackage{amsfonts}       
\usepackage{nicefrac}       
\usepackage{microtype}      

\usepackage{amsmath,amsfonts,amssymb,graphicx}
\usepackage{natbib}
\usepackage{comment}

\title{Integration of Leaky-Integrate-and-Fire-Neurons in Deep Learning Architectures}
\renewcommand{\undertitle}{Running title: LIF units in Deep Learning}

\author{
  Richard C.~Gerum \\
  Biophysics Group, Department of Physics\\
  Friedrich Alexander University Erlangen-Nürnberg (FAU), Germany \\
  \And
  Achim Schilling \\
  Experimental Otolaryngology,\\ Neuroscience Lab, University Hospital Erlangen, Germany \\
  Cognitive Computational Neuroscience Group\\at the Chair of English Philology and Linguistics, \\Friedrich-Alexander University Erlangen-Nürnberg (FAU), Germany\\
}

\renewcommand{\cite}{\citep}

\begin{document}

\maketitle

\noindent\textbf{Corresponding author:} \\
Dr. Achim Schilling  \\ 
Neuroscience Group \\
Experimental Otolaryngology \\
Friedrich-Alexander University of Erlangen-N\"urnberg \\
Waldstrasse 1 \\
91054 Erlangen, Germany \\
Phone:  +49 9131 8543853 \\
E-Mail: achim.schilling@uk-erlangen.de \\ \\

\newpage

\begin{abstract}
Up to now, modern Machine Learning is mainly based on fitting high dimensional functions to enormous data sets, taking advantage of huge hardware resources.
We show that biologically inspired neuron models such as the Leaky-Integrate-and-Fire (LIF) neurons provide novel and efficient ways of information encoding. They can be integrated in Machine Learning models, and are a potential target to improve Machine Learning performance.

Thus, we derived simple update-rules for the LIF units from the differential equations, which are easy to numerically integrate. We apply a novel approach to train the LIF units supervisedly via backpropagation, by assigning a constant value to the derivative of the neuron activation function exclusively for the backpropagation step. This simple mathematical trick helps to distribute the error between the neurons of the pre-connected layer. We apply our method to the IRIS blossoms image data set and show that the training technique can be used to train LIF neurons on image classification tasks. Furthermore, we show how to integrate our method in the KERAS (tensorflow) framework and efficiently run it on GPUs. To generate a deeper understanding of the mechanisms during training we developed interactive illustrations, which we provide online.

With this study we want to contribute to the current efforts to enhance Machine Intelligence by integrating principles from biology.

\end{abstract}

\newpage

\section*{Introduction}
The interest in „Neuroscience inspired AI“ has rapidly grown over the last years \cite{hassabis2017neuroscience}. There are major reasons for this development.
Although traditional Machine Learning algorithms have been massively improved by the collection of huge data sets \cite{russakovsky2015imagenet} and the development of modern hardware components \cite{steinkraus2005using, sheng2017distributed}, certain issues remain unsolved by these algorithms.
Up to now, these algorithms are —in contrast to our brain— highly specialized on a given task. We were not yet able to develop algorithms with general intelligence \cite{shevlin2019limits, pontes2019towards}.
Our nervous system has the ability to perform sensory tasks with enormous precision, such as the detection of very low stimuli in the eye \cite{field2019temporal, rieke1998single} or very small pressure differences in the ear and on the other hand is able to process and understand complex story plots \cite{mar2004neuropsychology,tenenbaum2006theory}.
Thus, we do not need huge hardware components but are limited to approximately $10^{11}$ neurons \cite{herculano2009human}, which perform these tasks in a very efficient way.
 
Information can be processed faster and more efficiently in the brain, as spiking neural networks can encode data in different spatio-temporal patterns \cite{thorpe2001spike, perkel1968neural, krauss2018statistical, gross1999origins}.
Thus, the brain does not simply count spikes (rate codes) but exploits the temporal dynamics of these spikes \cite{koopman2003dynamic} and uses spontaneous spiking and neural noise to enhance sensory processing \cite{schilling2020intrinsic, krauss2017adaptive, krauss2016stochastic}. 

Different biological inspired neuron models have been developed such as the Hodgkin-Huxley or the Fitzhugh-Nagumo neuron \cite{hodgkin1952quantitative, izhikevich2006fitzhugh}, but they are rarely integrated in Machine Learning applications.

Biologically inspired Leaky-Integrate-and-Fire (LIF) neurons \cite{burkitt2006review} are an interesting target for
Machine Learning models to potentially improve performance and increase interpretability on the one hand ("Neuroscience inspired AI" \cite{hassabis2017neuroscience}) and to create models for biology on the other hand (“Cognitive Computational Neuroscience” \cite{kriegeskorte2018cognitive}).

Due to these properties of spiking networks and especially LIF units much effort has be undertaken to train LIF neuron networks supervisedly via backpropagation.

In 2000 an interesting approach for training LIF networks via backpropagation was introduced, named SpikeProp, which was several times improved and further developed \cite{bohte2000spikeprop, schrauwen2004extending, schrauwen2004improving}. The basic idea of the algorithm is that not spiking patterns are learned, but desired spike times are defined as learning targets. 
In other approaches the input encoding is restricted to only one spike per neuron \cite{kheradpisheh2019s4nn}, meaning that the information is encoded in the time to first spike. 
Furthermore, some researchers use approximated derivatives in the gradient descent method to account for the discontinuities of the LIF neurons \cite{wu2018spatio, lee2020enabling}. 
Furthermore, some studies show that Long-Short-Term-Memory (LSTM) units \cite{hochreiter1997long} can also be trained, so that they behave like spiking neurons \cite{koopman2003dynamic, pozzi2018gating, rezaabad2020long}. This insight helps to use existing pipelines to train spiking neural networks. In this context, it has to be stated that in recent studies artificial spiking neural networks are more and more often compared to biological systems to understand and rebuild them \cite{kim2019simple}. Additionally, the spiking neural networks are an interesting target for energy efficient computation on small computer chips \cite{lee2020enabling}.
 
We here introduce a very direct way to train LIF neurons using backpropagation. Thus, we manually fix the derivative of the LIF neuron only within the backpropagation step and thus the error contributing to a certain spike can be distributed over the neurons of the pre-connected layer.
The study is structured as follows. We firstly illustrate and explain the function of LIF neurons. In a second step, we introduce our method to train LIF units with backpropagation. We prove the validity of our approach by the application of the method in hybrid neural networks trained on an image classification task using the IRIS data set \cite{nilsback2008automated}. The study comes along with interactive versions of most of the figures, which help the reader to gain a better understanding of the mechanisms within spiking neural networks.

\section*{Methods}
All simulations were run on a standard Desktop PC equipped with a Nvidia TitanXp Gpu device. The simulations were written in Python using Keras \cite{chollet2018deep} and Tensorflow \cite{abadi2016tensorflow} for Machine Learning and Numpy \cite{walt2011numpy} for further evaluations and interfaces. The visualization of the data was done in Javascript using the D$^3$-library \cite{bostock2011d3} and in Python using Matplotlib \cite{hunter2007matplotlib} and Pylustrator \cite{gerum2020pylustrator}. Thus, we provide interactive plots in an open github-repository. This interactive plots should help to gain a deeper understanding of the mechanisms described in the paper. A link to the repository is provided in the figure captions.

\section*{Results}
\subsection*{Leaky Integrate and Fire Neurons (LIF)}
\begin{figure}[!ht]
\centering
\includegraphics[]{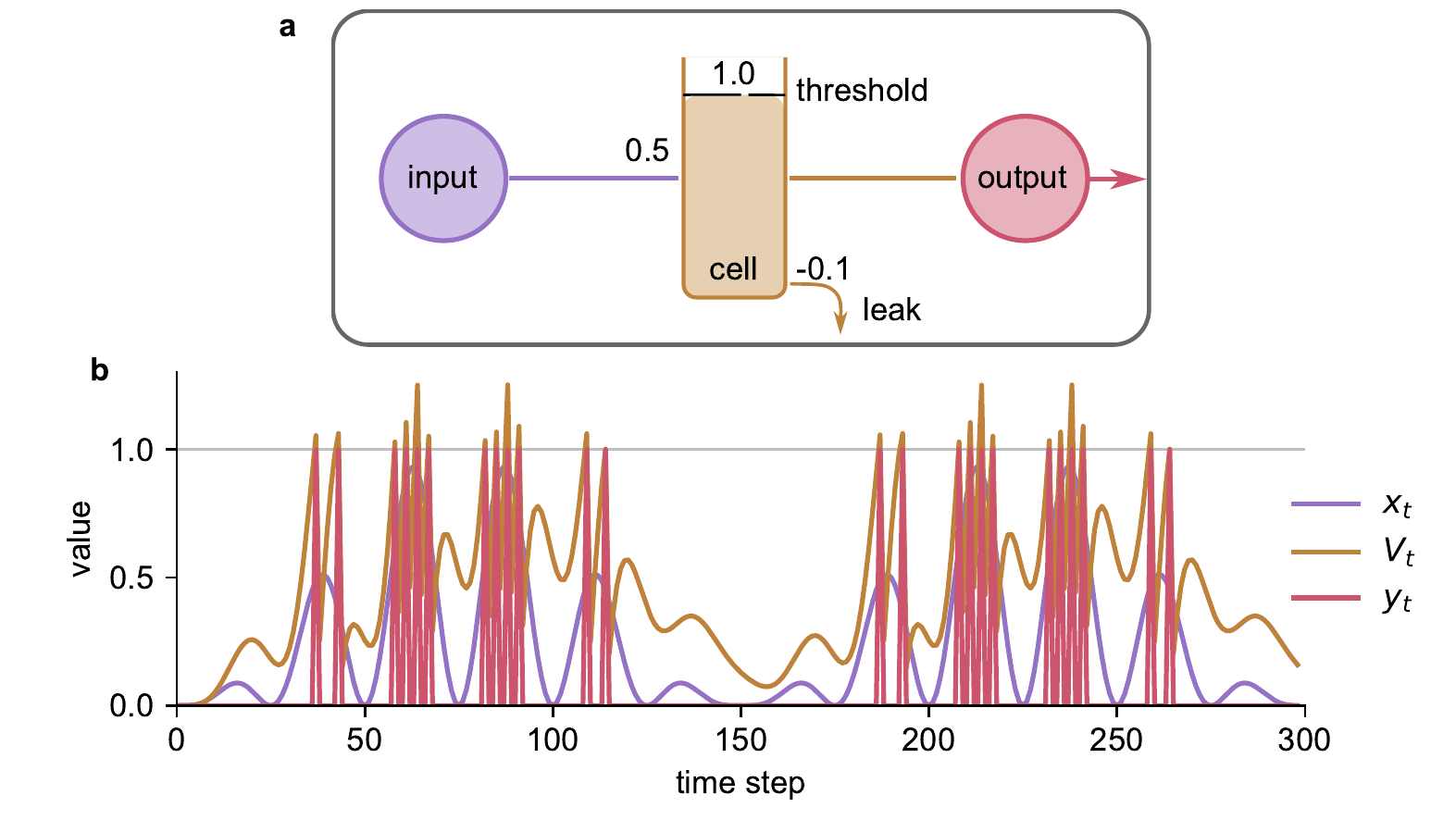}
\caption{
\textbf{The response of the leaky integrate and fire neurons.} \textbf{a}, illustration of the data flow in a LIF unit. \textbf{b}, the response of the LIF unit to a given input signal. The cell ($V_t$, orange) integrates the input ($x_t$, purple) until the internal state
exceeds the threshold (gray line). Then it outputs a spike ($y_t$, red). The leak term lets the cell state decay over time. With the parameters: 
$w_\mathrm{input} = 0.5$, $w_\mathrm{leak} = 0.1$,
$V_\mathrm{thresh} = 1.0$.
For interactive version see: \url{https://rgerum.github.io/paper_spiking_machine_intelligence/\#lif_unit}
 }
\label{fig:Lif_Spike}
\end{figure}

As described above the Leaky-Integrate-and-Fire (LIF) neuron model is a simple spiking neuron model based on one single differential equation.
The idea is that the neuron sums up all input currents, increases its' membrane potential and produces a spike if a certain threshold is reached (see Fig.\ \ref{fig:Lif_Spike}).
The leak term causes a continuous decrease of the membrane potential and thus prevents long-range correlations.
 
The leaky integrate and fire (LIF) neuron's \cite{koch1998methods} membrane potential $V_\mathrm{m}$ is described by the following differential equation:
\begin{align}
I(t) - \frac{V_\mathrm{m}(t)}{R_\mathrm{m}} = C_\mathrm{m} \cdot \dot V_\mathrm{m}(t)
\end{align}
(with the input current $I(t)$, the membrane resistance $R_\mathrm{m}$, and
the membrane capacity $C_\mathrm{m}$.)
 
When the membrane potential exceeds a threshold
 $V_\mathrm{th}$, a spike in form of a delta function $\delta(t)$ is emitted and the membrane potential is
reset to 0.
To simulate the response of a LIF neuron, this differential equation has to be integrated. The standard integration method is the Euler integration \cite{atkinson1989introduction}, which we used for this approach.
First, the differential equation is solved for $\dot V_\mathrm{m}(t)$
\begin{align}
\dot V_\mathrm{m}(t) = \frac{I(t)}{C_\mathrm{m}} - \frac{V_\mathrm{m}(t)}{R_\mathrm{m}C_\mathrm{m}}
\end{align}
 
This differential equation can be reformulated in a recursive manner (for one Euler step).
\begin{align}
V_{t+1} &= V_t + \left(C_\mathrm{m}^{-1} \cdot x_t - V_t \cdot R_\mathrm{m}^{-1}C_\mathrm{m}^{-1}\right)\cdot \Delta t
\end{align}
(with the time step delta $\Delta t$, and the $I(t)$ now renamed $x_t$.)
We extend the update equation to include the spiking when the threshold has been reached and the resetting of $V_{t+1}$ after the spike.
\begin{align}
\tilde V_{t+1} &= V_t + \left(C_\mathrm{m}^{-1} \cdot x_t - V_t \cdot R_\mathrm{m}^{-1}C_\mathrm{m}^{-1}\right)\cdot \Delta t&&\\
y_{t+1} &= \Theta(\tilde V_{t+1} - V_\mathrm{thresh}) && \mathsf{spiking}\\
V_{t+1} &= V_\mathrm{m} \cdot \Theta(-\tilde V_{t+1} + V_\mathrm{thresh}) && \mathsf{resetting}
\end{align}
Where $\Theta(x)$ is the Heaviside step function.
The parameters can be renamed as follows:
\begin{align}
C_\mathrm{m}^{-1} \Delta t &= w_\mathrm{input}\\
R_\mathrm{m}^{-1}C_\mathrm{m}^{-1}\cdot \Delta t &= w_\mathrm{leak}
\end{align}
 
Without loss of generality, $V_\mathrm{thresh}$ can be fixed to 1, as the scaling can be absorbed in $w_\mathrm{input}$.
The update rule of the LIF unit can be summarized as follows:
\begin{align}
V_{t} &= w_\mathrm{input} \cdot x_t + (1 - w_\mathrm{leak}) \cdot V_\mathrm{t-1} \cdot \Theta(V_\mathrm{thresh}-V_\mathrm{t-1})\\
y_{t} &= \Theta(V_{t} - V_\mathrm{thresh})
\label{eq:recursive}
\end{align}
These equations can e.g. be used to analytically calculate the firing rates $r$ of the LIF neurons (for calculation see Suppl. Fig. \ref{fig:Spike_Rate}).
The firing rates are an important property for many Machine Learning algorithms \cite{dominguez2016multilayer}.

\begin{align}
r = 1/\mathrm{ceil}\left(\frac{\ln\left(1- \frac{V_\mathrm{thresh}}{I} \cdot \frac{w_\mathrm{leak}}{w_\mathrm{input}}\right)}{\ln{(1-w_\mathrm{leak})}} - 1\right)
\end{align}
Here, $\mathrm{ceil}(x)$, denotes the ceiling of a number, i.e.\ rounding up to the closest integer.

\subsection*{Deep Learning with LIF Neurons}
We show how the LIF neurons can be integrated in standard Machine learning applications for image classification.

\subsubsection*{LIF Neurons and Multidimensional Data}
\begin{figure}[!ht]
\centering
\includegraphics{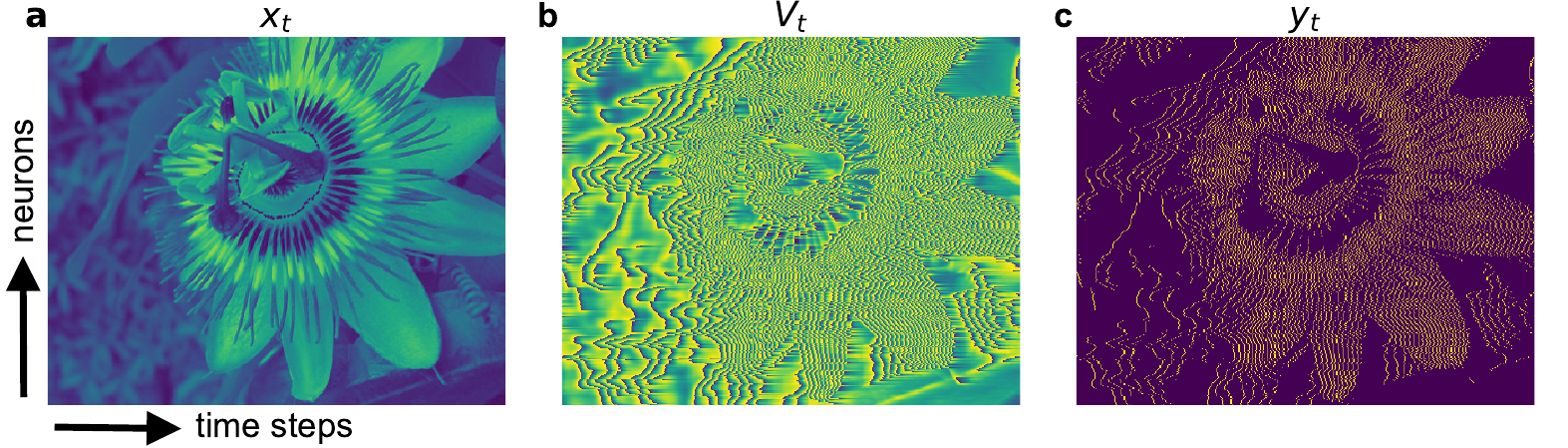}
\caption{\textbf{Image processed column-wise by a LIF layer.} The figure shows the input (a), internal state (b), and the output (c) of the LIF layer.
 For interactive version see: \url{https://rgerum.github.io/paper_spiking_machine_intelligence/\#lif_image_processing}
}
\label{fig:Spike_Blossom}
\end{figure}

When the LIF neurons are applied to multi-dimensional data such as an image an efficient representation is needed to optimize LIF neurons for our standard hardware and software architectures. The image, which we feed to the LIF units has $N$ rows and $M$ columns ($N\times M$).
However, we regard the image as serial data set, where the time axis corresponds to the x-axis of the image. Thus, the input of the LIF neurons consists of an
$N$-dimensional vector for each of the $M$ time steps. Therefore, also $V_t$ and $y_t$ are $N$-dimensional vectors.
Thus, when an image is fed to a LIF unit as described above (input image see Fig.\ \ref{fig:Spike_Blossom}a) it is transformed to "voltage fluctuations" in the LIF unit (see Fig.\ \ref{fig:Spike_Blossom}b) and output spiking patterns (see Fig.\ \ref{fig:Spike_Blossom}c).

The model can be extended by the use of several LIF units with different $w_{\mathrm{input}}$ and $w_{\mathrm{leak}}$ (see eq. \ref{eq:recursive} ). This would make $w_{\mathrm{input}}$ and $w_{\mathrm{leak}}$ a vector instead of a scalar and the scalar product would transform into a tensor product. This is an efficient representation, which can easily be optimized for and run on GPUs.
 
\subsubsection*{Calculation of the Gradient}
The standard method to supervisedly train neural networks on a classification task is to minimize a loss function $L(y_{\mathrm{out}}, y_{\mathrm{desired}})$.
It is a measure of the dissimilarity between the desired output and the output of the neural network $y_{\mathrm{out}}$ calculated by forward propagation.
For example in a classification task with a softmax output, the loss function usually is the cross-entropy.
This loss function is minimized using a gradient descent algorithm.
The gradient descent works by adding the negative gradient multiplied with the learning rate $\gamma$ to the weights, which have to be optimized.
\begin{align}
\Delta W = - \gamma\cdot \frac{\mathrm{d}L}{\mathrm{d}W}
\end{align}

\begin{figure}[ht]
\begin{center}
\includegraphics{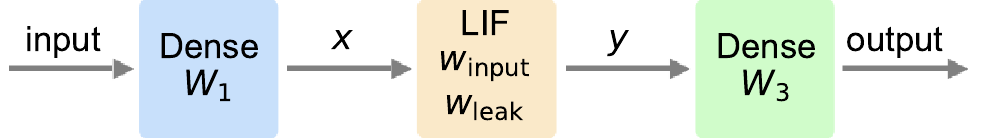}
\end{center}
\caption{\textbf{Example network architecture}\newline Example neural network architecture used to illustrate the gradient descent algorithm.}
\label{fig:example_architecture}
\end{figure}

To illustrate the calculation of the gradient, we use an example architecture, consisting of two fully-connected layers and a LIF layer in between (see Fig. \ref{fig:example_architecture}).

To calculate the update of the weights $W_1$, we have to calculate the gradient $\frac{\mathrm{d}L}{\mathrm{d}W_1}$.
\begin{align}
\frac{\mathrm{d}L}{\mathrm{d}W_1} = \frac{\partial L}{\partial y_\mathrm{out}} \cdot{\frac{\partial y_{\mathrm{out}}}{\partial y}}\cdot{\frac{\partial y}{\partial x}}\cdot\frac{\mathrm{d} x}{\mathrm{d} W_1}
\end{align}

The term ${\frac{\partial y}{\partial x}}$ is the derivative of the LIF output as a function of the LIF input.
If this derivative is 0 the weights of the first layer $W_1$ cannot be trained.
 
\begin{align}
V_{t} &= w_\mathrm{input} \cdot x_t + (1 - w_\mathrm{leak}) \cdot V_\mathrm{t-1} \cdot \Theta_2(V_\mathrm{thresh}-V_\mathrm{t-1})\\
y_{t} &= \Theta_1(V_{t} - V_\mathrm{thresh})
\label{eq: differential_2}
\end{align}

However, this is exactly the case, as the LIF equations contain two $\Theta$ functions. To better reference them, we call the $\Theta$ function for generation the output spike $\Theta_1$ and the $\Theta$ function for resetting the membrane potential $\Theta_2$.
As the $\Theta$ function is a completely flat function (except at 0), its' gradient is 0 at all points, therefore reducing all gradients to 0. Thus, no gradient can enter or pass the LIF cell.
One possibility to overcome this problem would be to smooth the $\Theta$ functions. But a more elegant solution, that does not affect the forward pass, is to just redefine the gradient of the LIF unit. The derivative with respect to the inputs can be written as follows:
\begin{align}
\frac{\partial y_t}{\partial x_t} &= \frac{\partial y_t}{\partial V_t} \frac{\partial V_t}{\partial x_t} = \Theta^\prime_1(V_t - V_\mathrm{thresh}) \cdot w_\mathrm{input}\\
\frac{\partial y_t}{\partial x_{t-1}} &= \frac{\partial y_t}{\partial V_t} \frac{\partial V_t}{\partial V_{t-1}} \frac{\partial V_{t-1}}{\partial x_{t-1}}\\
&= \Theta^\prime_1(V_t - V_\mathrm{thresh}) \cdot (1- w_\mathrm{leak}) \cdot \left[ \Theta_2(V_\mathrm{thresh} - V_{t-1}) + V_{t-1} \cdot \Theta^\prime_2(V_\mathrm{thresh}-V_{t-1}) \right] \cdot w_\mathrm{input}
\end{align}

If we define $\Theta^\prime_2(x)=0$, then the expression for an arbitrary derivative for a past $x$ is:
\begin{gather}
\frac{\partial y_t}{\partial x_{t-n}} = \Theta^\prime_1(V_t - V_\mathrm{thresh}) \cdot w_\mathrm{input}  (1- w_\mathrm{leak})^{n} \prod_{i=1}^{n} \Theta_2(V_\mathrm{thresh} - V_{t-1})
\end{gather}

One can see that the crucial part here is the function $\Theta_1$ that prevents any gradient to pass. Therefore, we redefine the gradient of $\Theta_1$ to be 1 and keep the gradient of $\Theta_2$ as 0. 
This procedure is only a small change to the equations, which only affects the backpropagation step and leaves the forward propagation untouched.

\begin{figure}
\centering
\includegraphics{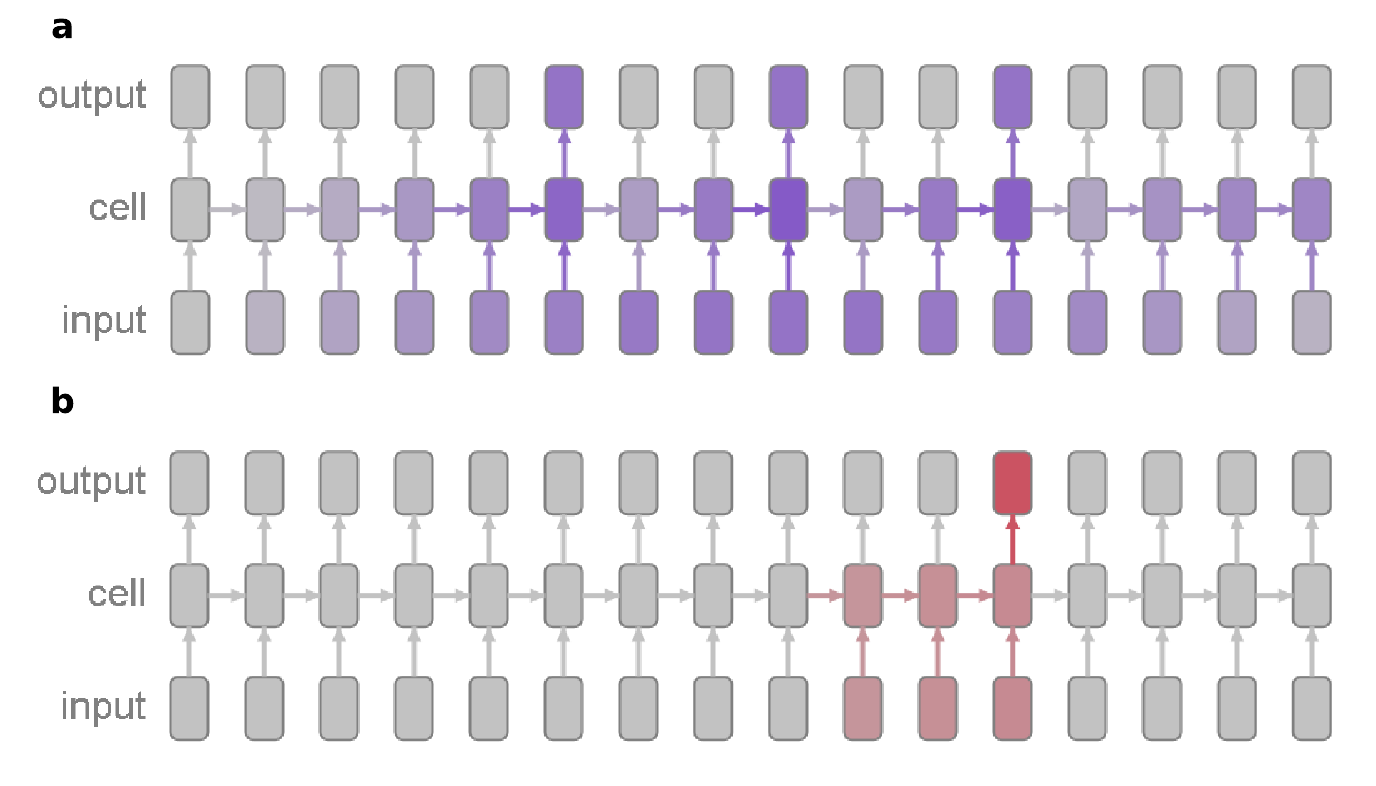}
\caption{\textbf{The response and gradient of a LIF cell.} Each column represents one time step. The color saturation denotes the value of the cells which results from the inputs \textbf{a}. \textbf{b}, the propagated error from the red time step.
The derivatives are defined as follows: $\Theta^\prime_1(x) = 1$, $\Theta^\prime_2(x) = 0$.
For interactive version see: \url{https://rgerum.github.io/paper_spiking_machine_intelligence/\#lif_backpropagation}
 }
 \label{fig:gradient_flow}
\end{figure}

The gradient enters the cell (as $\Theta^\prime_1=1$) and propagates to all input units that contributed to the spike (see Fig. \ref{fig:gradient_flow}) but does not penetrate to inputs that contributed to the previous spike (as $\Theta^\prime_2=0$).
We have shown that our gradient definition allows errors to pass the LIF neurons to pre-connected layers.
For the backpropagation procedure the gradients with respect to the LIF parameters is also needed, if they are supposed to be trainable ($w_{\mathrm{input}}$, $w_{\mathrm{leak}}$, for complete gradients see Supplements).

In the next step, the LIF unit implementation described above is embedded in a hybrid neural network out of LSTM layers and a softmax layer. This hybrid neural network is applied to an image classification task, where 10 different flower species should be identified.
We are aware of the fact that the LSTM units add further complex effects to the model, but we use them as they are a simple method to integrate the spikes and transform them into a class label.
The used data set is a sub-data set of the 102 category flower data set \cite{nilsback2008automated} with only 10 categories.
Thus, the LIF units should pre-process and compress the images of the different blossoms.

\subsection*{LIF Units for Image Classification}

\subsubsection*{Network Architecture 1}
The classification task on different blossoms is based on the 10 most occurring flower species (see Suppl. Fig. \ref{fig:Dataset}) of the 102 category flower data set \cite{nilsback2008automated}.
The used network consists of one LIF layer with 3 different LIF units types (3x2=6 trainable parameters) compressing the colored images of 500x400 pixels.

The three LIF unit types each get exactly one color channel of the input images. In each time step each LIF unit type receives one column of one color channel of the image as input and returns a spike vector still representing the same color channel.
Thus, the view that the LIF layer consists of 1200 individual LIF neurons of 3 different sorts (in analogy to network architecture 2) with 1D spike train output is equivalent, although for programming reasons the tensor notation was used in the tensorflow \cite{tensorflow2015-whitepaper} implementation.
 
Thus, the 8 bit images are compressed by a factor of 8 as each 8 bit integer is replaced by a boolean number (spike, no spike). The compressed spike data is fed to an LSTM layer (30 units) connected to a fully connected output layer with softmax activation (10 units) (see Tab.\ \ref{tab:net1}). As loss function the categorical cross-entropy is used. The parameters $w_{\mathrm{input}}$ and $w_{\mathrm{leak}}$ of the LIF units, as well as the LSTM and softmax parameters are trained via backpropagation.

\begin{table}
\caption{\textbf{Network architecture 1}}
\label{tab:net1}
\centering
\begin{tabular}{lll}
Layer (type) & Output Shape & Parameters \# \\
\hline
LIF-Layer & (None, 400, 500, 3) & 6 \\
Reshape & (None, 500, 1200) & 0 \\
LSTM Layer & (None, 500, 30) & 147720 \\
Dropout & (None, 500, 30) & 0 \\
Time distributed Dense & (None, 500, 30) & 930 \\
Dropout & (None, 500, 30) & 0 \\
Softmax & (None, 500, 10) & 310 \\
\hline
\end{tabular}
\end{table}

\begin{figure}
\centering
\includegraphics[]{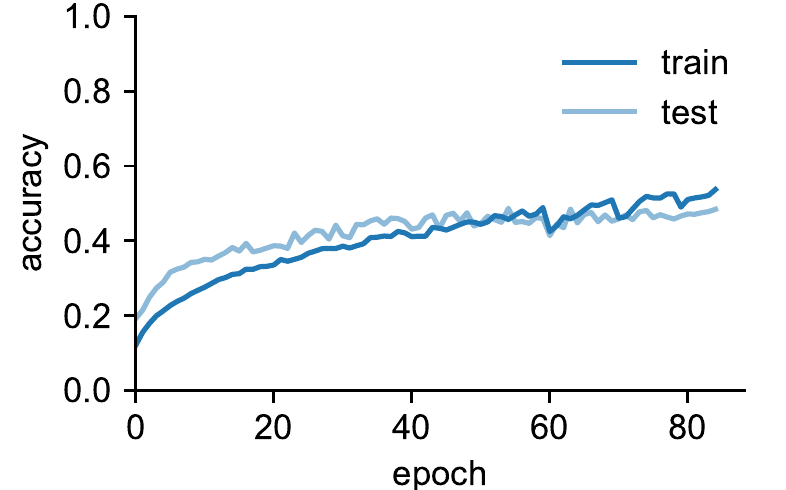}
\caption{
\textbf{Accuracy of training a network with a LIF layer.} The training accuracy is shown in dark blue and the test accuracy in light blue.}
 \label{fig:training1}
\end{figure}

The training procedure is stopped after 30 epochs of no improvement of the test accuracy ("early stopping").
The classification accuracy for one image (see fig.\ \ref{fig:training1}) is defined as the average probability value of the correct label during the image presentation, which is a very conservative estimator for the accuracy.
The overall test accuracy (all test images) achieves a value of over 40 $\%$, whereas the chance accuracy is 10$\%$ (10 categories). This proves that the LIF neurons can be trained via backpropagation so that they operate in a sophisticated parameter range.

Note that the data set does not allow the neural network to train on trivial features such as the color of the flower as the data set contains different color variants of the same flower species.
\begin{figure}
\centering
\includegraphics{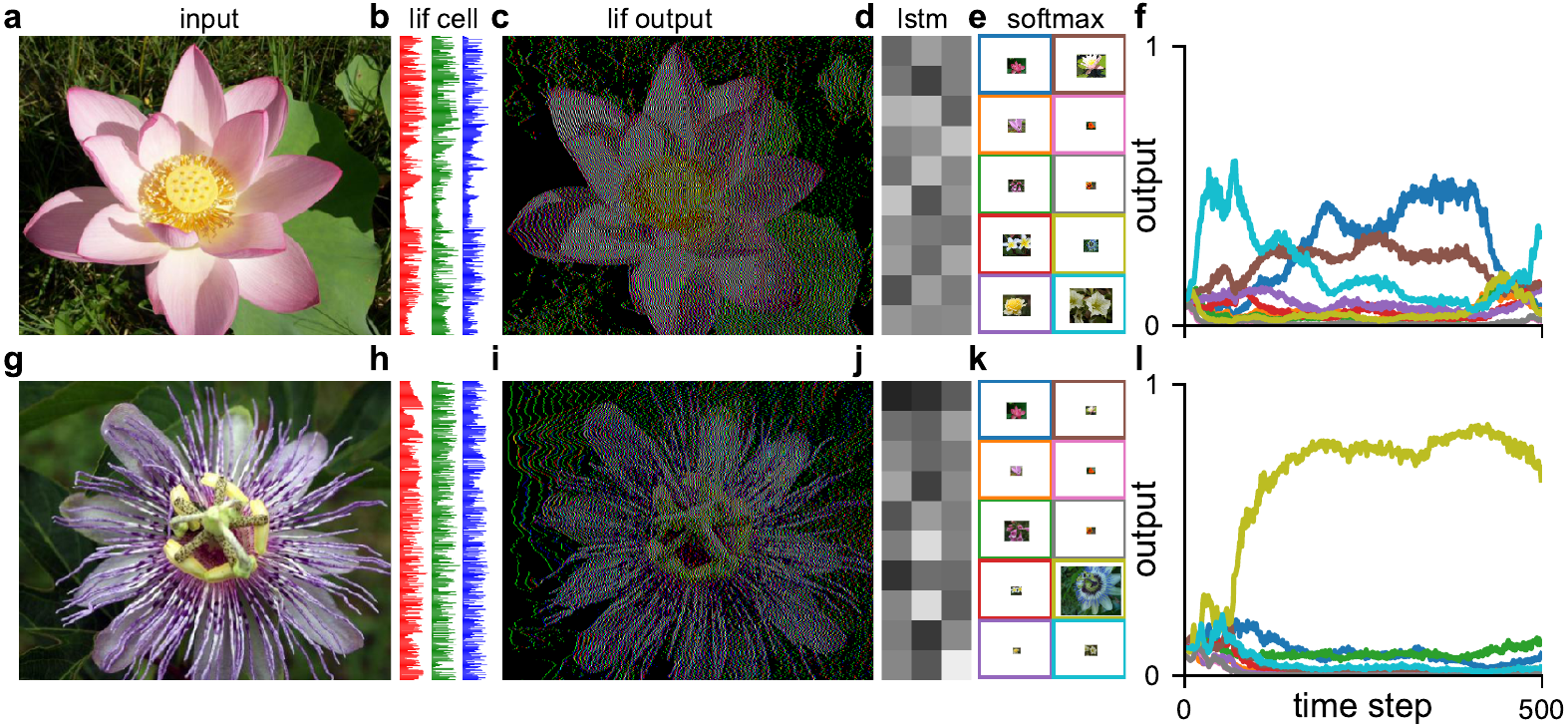}
\caption{
\textbf{Spiking network processes images.} The network takes images as input (\textbf{a}, \textbf{g}) and applies a LIF layer (\textbf{b}, \textbf{h}) to generate spike trains for each row and color channel (\textbf{c}, \textbf{i}). 
These spike trains are fed into an LSTM layer (\textbf{d}, \textbf{j}) which is followed by a softmax layer (\textbf{e}, \textbf{k}). The softmax layer predicts the category of the image (\textbf{f}, \textbf{l}). The three different colorbars (lif cell) represent the internal state of the LIF units for the three different color channels. The spike data produced by this LIF layer is shown under the heading lif output.
The activation of the LSTM layer is shown as color map (lstm). The category probability calculated through the softmax layer is represented by the size of the category images (softmax).
For interactive version see: \url{https://rgerum.github.io/paper_spiking_machine_intelligence/\#lif_network1}
}
\label{fig:network1_vis}
\end{figure}
The LIF units compress the image, nevertheless the shape of the flowers can still be seen in the spike patterns (see Fig. \ref{fig:network1_vis}).
The here described model proves that spiking layers can be trained for a classification task.

\subsubsection*{Network Architecture 2}
\begin{table}
\caption{\textbf{Network architecture 2}}
\label{tab:net2}
\centering
\begin{tabular}{lll}
Layer (type) & Output Shape & Parameters \# \\
\hline
Reshape & (None, 500, 1200) & 0 \\
TimeDistributed Dense & (None, 500, 1200) & 1441200 \\
LIF-Layer & (None, 500, 1200) & 2400 \\
LSTM Layer & (None, 500, 30) & 147720 \\
Dropout & (None, 500, 30) & 0 \\
Time distributed Dense & (None, 500, 30) & 930 \\
Dropout & (None, 500, 30) & 0 \\
Softmax & (None, 500, 10) & 310 \\
\hline
\end{tabular}
\end{table}

We further provide evidence that a classification network can also be trained, when there is a fully-connected layer pre-connected to the LIF layer.
Here, the LIF layer consists of 1200 LIF units, generating output spike trains (1D boolean scalar spike train). Each LIF unit receives a weighted sum of $3\cdot400$ values as input (3 color channels and 400 as the images consist of 400 rows).
The x-coordinate (500 pixels width of the image) is the time axis.
 
In contrast to the architecture above (network 1), with only 6 trainable parameters except LSTM and softmax layer (3 LIF neuron types, 1200 LIF neurons), this network has 1,441,200 trainable parameters for the pre-connected fully-connected layer
and 2,400 trainable parameters ($w_\mathrm{input}$, $w_\mathrm{leak}$) for the 1,200 individual LIF units (see Tab.\ \ref{tab:net2}).
The fact that the gradient can pass the LIF units can be seen when analyzing the accuracy as a function of the epochs (learning curve).
The accuracy is higher for the LIF units with the activation function gradient ($\Theta_1^{\prime} =1$) set to one (blue curves). This is true for training as well as test accuracy.
\begin{figure}
\centering
\includegraphics[]{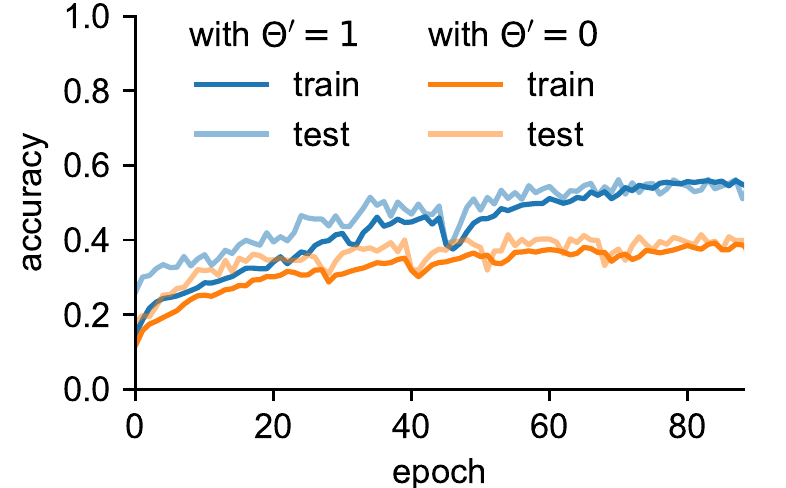}
\caption{
\textbf{Accuracy of training a network with included LIF layer.} The orange curves show the accuracy during training without manually setting the gradient of $\Theta_1^{\prime}=1$ (disappearing gradient network, dark orange: training accuracy, light orange: test accuracy).
The gradient cannot pass the LIF units. The blue curves in contrast show that the algorithm is able to train the LIF layer as well as the pre-connected dense layer (dark blue: training accuracy, light blue: test accuracy).}
\label{fig:training2}
\end{figure}

\begin{figure}[htb]
\centering
\includegraphics{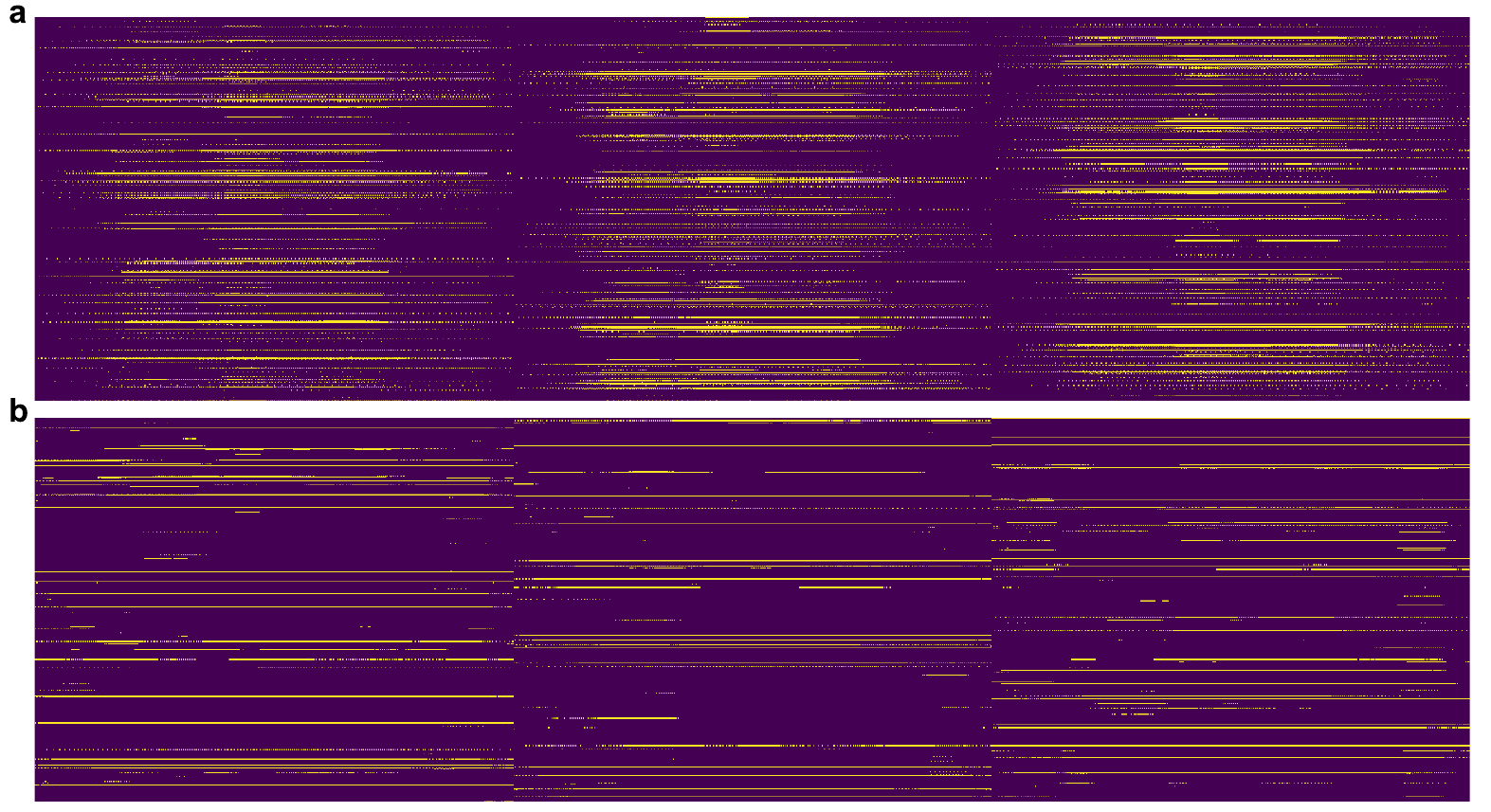}
\caption{
\textbf{Training of a network with LIF units.} 
\textbf{a} the spiking patterns of the LIF units before training (epoch 0, each line in the three blocks represents one of the 1200 neurons, the x-axis is the 500 time points, yellow represents a spike, purple represents no spike). \textbf{b} the spiking patterns of the LIF units after training (epoch 172).
The input of one LIF unit in each time step is a weighted sum of the rows of the image (each image has 400 rows and 3 color channels, input: weighted sum of 1200 values). It can be seen that the output spike patterns change during training (\textbf{a}, epoch 0, \textbf{b} epoch 172) and that the spike density is reduced.
Thus, the network develops a sparse coding of the input image.
For interactive version see: \url{https://rgerum.github.io/paper_spiking_machine_intelligence/\#lif_network2}
}
\label{fig:LiF_net2_output}
\end{figure}

The test accuracy for the network, where the gradient can pass the LIF layer, is increased by more than 10 $\%$ compared to the disappearing gradient network (orange curve) and raises up to a value of approximately 55 $\%$ (see Fig.\ \ref{fig:training2}).
Nevertheless, the accuracy of the disappearing gradient network is not at chance level, as the random connections lead to usable features for the higher layers (LSTM layers), an effect that was shown in biology as well as in computer science (see e.g.\ \cite{dasgupta2017neural}).
In the following, we show that the output of the LIF layer significantly changes over the training epochs. Thus, the time course of the output of a LIF unit for one certain image is shown in Fig. \ref{fig:LiF_net2_output}.

The spike patterns clearly change during the training process and the spike occurrences decreases (see Fig.\ \ref{fig:LiF_net2_output}). This effect --called sparse-coding \cite{olshausen1997sparse}-- is correlated with a higher performance of the network and shows that the gradient can pass the LIF layer. Furthermore, sparse-coding of input stimuli is a basic principle in biological neural networks and here emerges automatically due to our training method.

\section*{Discussion}

\subsection*{Summary}

In this study, we have done an in-depth analysis of the maths behind, and the applicability of LIF neurons in Machine Learning.

We show that LIF units can be embedded in standard neural network models and can be trained with backpropagation. We developed a definition of the gradient, that allows the backpropagated error of a spike to be assigned to those inputs that contributed to this spike. This was done by fixing the derivatives of the activation functions ($\Theta_1^{\prime} = 1$, $\Theta_2^{\prime}=0$).
 
For our analysis we chose a complex image data set, which has in contrast to simpler data sets such as MNIST, more different frequencies, which translate into different spiking patterns.
Up to now, in most studies only simpler data sets were used (for review see \citet{tavanaei2019deep}). 

\subsection*{Limitations}
The aim of our study was to establish a simple idea how to train neural networks consisting of LIF neurons, although we are aware of the fact that we do not yet achieve state-of-the-art performance compared to other architectures optimized for image classification (see e.g.\ \cite{xia2017inception, feng2019flower, qin2019new}).
Additionally, to achieve higher performance the recurrent neural networks have to be optimized for the used hardware components \cite{bhuiyan2010acceleration}.

However, we are convinced that our method is a very direct and uncomplicated way to train LIF networks and thus is interesting to the Machine Learning as well as Neuroscience community. 
Furthermore, our method is easy to implement in existing Machine Learning frameworks such as Keras \cite{chollet2018deep} and thus can efficiently been run on GPU devices. The interactive visualizations help to further understand the computational mechanisms and are a step towards explainable AI. To gain a deeper understanding how Machine Learning algorithms work, i.e.\ to solve the black-box problem (also opacity debate), has currently become an important issue in AI research \cite{castelvecchi2016can, de2018algorithmic}.

The results of this study have the potential to provide novel insights in the function of biological neural networks (cf.\ e.g.\ \cite{schemmel2006implementing, jin2010modeling}).
Thus, we think that the application of analysis techniques developed for untrained or randomly connected neural networks such as stability analysis (Liapunov exponent) or motif distribution analysis (see e.g. \cite{bertschinger2004real, krauss2019analysis, krauss2019recurrence, krauss2019weight}), can be applied to the spiking neural networks trained with backpropagation to gain new insights in brain dynamics and function.

\subsection*{Concluding Remarks}
The implementation of biological principles in Machine Learning such as sparsity \cite{gerum2020sparsity} or spiking properties can help to improve the performance of Machine Learning algorithms.
Furthermore, we hypothesize that spiking neural networks unfold their full potential in tasks with serial data such as speech or music classification \cite{schilling2020intrinsic}.

\section*{Acknowledgements}
We thank Nvidia for the donation of two Titan Xp Gpu devices.

\bibliographystyle{apalike}
\bibliography{references}

\renewcommand\thefigure{S\arabic{figure}}    
\setcounter{figure}{0}

\section*{Supplementary Material}
\subsection*{Firing Rates}

We will analytically calculate the number of timesteps with a constant input $I$ which are necessary to provoke a spike.
At time $t_0$ we start with $V_{t_0}=w_{input}\cdot I$ and apply the update rule for LIF units as shown above:
\begin{align}
V_{t_n} &= w_\mathrm{input} \cdot I + (1-w_\mathrm{leak}) \cdot V_{t_n-1}
\end{align}

The implicit description of $V_{t_n}$ can be written in an explicit manner.
\begin{align}
V_{t_n} &= \sum_{i=0}^n (1-w_\mathrm{leak})^i \cdot w_\mathrm{input} \cdot I
\end{align}

The criterion for a spike is: $V_{t_n} \ge V_\mathrm{thresh}$.
Thus, the following inequation has to be solved:
\begin{align}
V_\mathrm{thresh} &\le  \sum_{i=0}^n (1-w_\mathrm{leak})^i \cdot w_\mathrm{input} \cdot I\\
&\le w_\mathrm{input} \cdot I \cdot \sum_{i=0}^n (1-w_\mathrm{leak})^i ~~~~ \mathrm{with} \sum_{i=0}^n a^i = \frac{a^{n + 1} - 1}{a - 1} \\
&\le w_\mathrm{input} \cdot I \cdot \frac{(1-w_\mathrm{leak})^{n+1}-1}{-w_\mathrm{leak}}\\
-\frac{V_\mathrm{thresh}}{I}\cdot \frac{w_\mathrm{leak}}{w_\mathrm{input}} + 1 &\ge (1-w_\mathrm{leak})^{n+1}\\
\ln\left(-\frac{V_\mathrm{thresh}}{I}\cdot \frac{w_\mathrm{leak}}{w_\mathrm{input}} + 1\right) &\ge (n+1) \cdot \ln(1-w_\mathrm{leak})\\
n &\ge \frac{\ln\left(1- \frac{V_\mathrm{thresh}}{I} \cdot \frac{w_\mathrm{leak}}{w_\mathrm{input}}\right)}{\ln{(1-w_\mathrm{leak})}} - 1\\
\end{align}

This results in the number of timesteps until the next spike event.
\begin{align}
n = \mathrm{ceil}\left(\frac{\ln\left(1- \frac{V_\mathrm{thresh}}{I} \cdot \frac{w_\mathrm{leak}}{w_\mathrm{input}}\right)}{\ln{(1-w_\mathrm{leak})}} - 1\right)
\end{align}

Note, for $w_\mathrm{leak}=0$ this simplifies to:
\begin{align}
n=\mathrm{ceil}\left(\frac{V_\mathrm{thresh}}{w_\mathrm{input}\cdot I}\right)
\end{align}

The equation shows that $n$ is a nonlinear function of $w_{\mathrm{leak}}$ and $w_{\mathrm{input}}$ and thus a simple downregulation of $w_{\mathrm{leak}}$ cannot be compensated by a downregulation of $w_{\mathrm{input}}$. On the other hand, $V_\mathrm{thresh}$ can be fixed to 1 without loss of generality as a change of $V_\mathrm{thresh}$ can be absorbed in $w_\mathrm{input}$.

The spiking only occurs at all if the input exceeds the input threshold:
\begin{align}
I_\mathrm{min} = \frac{V_\mathrm{thresh} \cdot w_\mathrm{leak}}{w_\mathrm{input}}
\end{align}

Smaller inputs lead to a divergence of the $n$ as the cell state decays faster than it is resupplied by the input.
The equation for the input threshold $I_\mathrm{min}$ shows that the input threshold is only non-zero if the neuron is "leaky", e.g. has a leak parameter $w_\mathrm{leak} \ne 0$.
 
\begin{figure}
\centering
\includegraphics[scale=0.7]{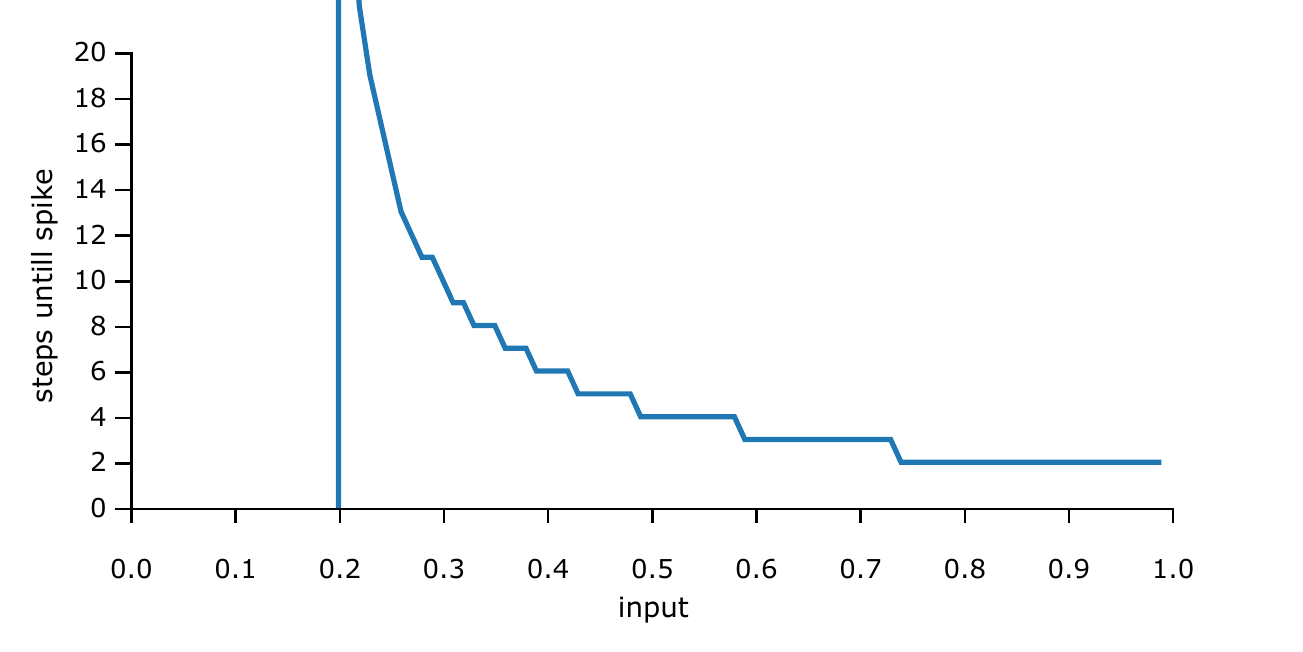}
\caption{\textbf{Average number of timesteps needed for spike (corresponds to inverse spike rate).} The curve (blue)
shows the steps it takes for the neuron to spike when it receives a constant input ($w_\mathrm{input}=0.5$, $w_\mathrm{leak} = 0.1$).}
\label{fig:Spike_Rate}
\end{figure}

The fact that the two parameters $w_{\mathrm{leak}}$ and $w_{\mathrm{input}}$ have different effects on the spike rate,
and cannot be trivially combined to one parameter, training of these two parameters can help to extract interesting features from the input data.

\subsection*{Limited image data set}
The neural networks used in this study were trained on the classification of 10 different flower species, which are a subset of 102 Oxford flower data set \cite{nilsback2008automated}.

\begin{figure}[!ht]
\centering
\includegraphics[scale=0.7]{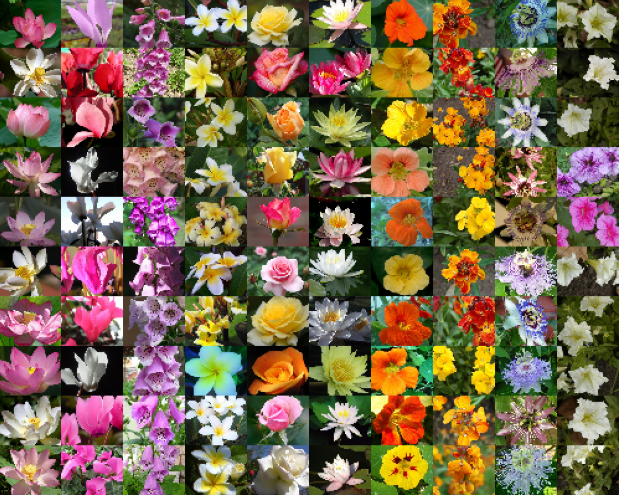}
\caption{\textbf{Exemplary images from the limited data set with 10 categories out of 102 Oxford flower data set}}
\label{fig:Dataset}
\end{figure}

\subsection*{Gradients}
The gradient with respect to $w_{\mathrm{input}}$:
\begin{gather}
\begin{array}{l}
\frac{\mathrm{d}y_t}{\mathrm{d}w_{\mathrm{input}}} =
\frac{\partial y_t}{\partial V_t}\left( \frac{\partial V_t}{\partial w_{\mathrm{input}}}+\frac{\partial V_t}{\partial V_{t-1}}\frac{\mathrm{d}V_{t-1}}{\mathrm{d}w_{\mathrm{input}}}\right)=\\
\frac{\partial y_t}{\partial V_t}\Big\{\frac{\partial V_t}{\partial w_{\mathrm{input}}}+\frac{\partial V_t}{\partial V_{t-1}} \left[\frac{\partial V_{t-1}}{\partial w_{\mathrm{input}}}+\frac{\partial V_{t-1}}{\partial V_{t-2}}\left(\frac{\partial V_{t-2}}{\partial w_{\mathrm{input}}}+\frac{\partial V_{t-2}}{\partial V_{t-3}}\frac{\mathrm{d} V_{t-3}}{\mathrm{d} w_{\mathrm{input}}} \right)\right]\Big\}=\\
\Theta_1^\prime(V_t-V_{\mathrm{thresh}})\cdot \Big[x_t + (1-w_{\mathrm{leak}})\Theta_2(V_{\mathrm{thresh}}-V_{t-1})x_{t-1}+\\
(1-w_{\mathrm{leak}})^2\Theta_2(V_{\mathrm{thresh}}-V_{t-1})\Theta_2(V_{\mathrm{thresh}}-V_{t-2})x_{t-2}+\\
(1-w_{\mathrm{leak}})^3\Theta_2(V_{\mathrm{thresh}}-V_{t-1})\Theta_2(V_{\mathrm{thresh}}-V_{t-2})\Theta_2(V_{\mathrm{thresh}}-V_{t-2})\frac{\mathrm{d} V_{t-3}}{\mathrm{d} w_{\mathrm{input}}}\Big]=\\
\Theta_1^\prime(V_t-V_{\mathrm{thresh}})\cdot\Big[x_t+\sum_{n=1}^{N}{x_{t-n}(1-w_{\mathrm{leak}})^n \cdot \prod_{i=1}^{n}{\Theta_2(V_{\mathrm{thresh}}-V_{t-i})}}\Big]
\end{array}
\end{gather}

The gradient with respect to $w_{\mathrm{leak}}$:
\begin{gather}
\begin{array}{l}
\frac{\mathrm{d}y_t}{\mathrm{d}w_{\mathrm{leak}}} =
\frac{\partial y_t}{\partial w_{\mathrm{leak}}}\frac{\mathrm{d}V_t}{\mathrm{d}w_{\mathrm{leak}}}=\\
=\Theta_1^\prime(V_t-V_{\mathrm{thresh}})\cdot\\ \Big[-V_{t-1}\Theta_2(V_{\mathrm{thresh}}-V_{t-1})-V_{t-2}(1-w_{\mathrm{leak}})\Theta_2(V_{\mathrm{thresh}}-V_{t-1})\Theta_2(V_{\mathrm{thresh}}-V_{t-2})-\\
V_{t-3}(1-w_{\mathrm{leak}})^2\Theta_2(V_{\mathrm{thresh}}-V_{t-1})\Theta_2(V_{\mathrm{thresh}}-V_{t-2})\Theta_2(V_{\mathrm{thresh}}-V_{t-3})+\\
(1-w_{\mathrm{leak}})^3\Theta_2(V_{\mathrm{thresh}}-V_{t-1})\Theta_2(V_{\mathrm{thresh}}-V_{t-2})\Theta_2(V_{\mathrm{thresh}}-V_{t-3})\frac{\mathrm{d}V_{t-3}}{\mathrm{d}w_{\mathrm{leak}}}\Big]=\\
\Theta_1^\prime(V_t-V_{\mathrm{thresh}})\cdot\Big[-\sum_{n_1}^N{V_{t-n}(1-w_{\mathrm{leak}})^{n-1}}\cdot \prod_{i=1}^n{\Theta_2(V_{\mathrm{thresh}}-V_{t-i})}\Big]
\end{array}
\end{gather} 

\subsection*{Keras Implementation}
These gradient definitions can be implemented with the following code in tensorflow:
 
\begin{verbatim}
@tf.function
def lif_gradient(x, w_i, w_l, t_thresh=1):
    time_steps = x.shape[1]

    Vm = w_i * x[:, 0]
    states = tf.TensorArray(tf.float32, size=time_steps)

    for i in tf.range(time_steps):
        Vm = w_i * x[:, i] + (1 - w_l) * Vm * theta2(t_thresh - Vm)
        spike = theta1(Vm - t_thresh)
        states = states.write(i, spike)

    return tf.transpose(states.stack())
\end{verbatim}

With the theta functions defined as:
 
\begin{verbatim}
@tf.custom_gradient
def theta2(x):
    def grad(dy):
        return dy*0
    return tf.cast(x > 0, tf.float32), grad

@tf.custom_gradient
def theta1(x):
    def grad(dy):
        return dy*1
    return tf.cast(x > 0, tf.float32), grad
 \end{verbatim}

\end{document}